\pdfoutput=1

\documentclass[11pt]{article}

\usepackage{acl}

\usepackage{times}
\usepackage{latexsym}

\usepackage[T1]{fontenc}

\usepackage[utf8]{inputenc}

\usepackage{microtype}

\usepackage{inconsolata}

\usepackage{graphicx}

\usepackage{subfig}
\usepackage{amsmath,amssymb,amsfonts}
\usepackage{booktabs}
\usepackage{tabularx}
\usepackage{multicol}
\usepackage{array}
\usepackage{seqsplit}
\usepackage{tcolorbox}
\usepackage{seqsplit}
\usepackage{float}
\usepackage{fixltx2e}
\usepackage{pgffor} 
\usepackage{kotex}
\usepackage{listings}
\usepackage{xcolor} 
\usepackage{multirow}
\definecolor{LightGray}{gray}{0.9}
\usepackage[frozencache=true,cachedir=./minted-cache]{minted}
\definecolor{mildred}{RGB}{200, 50, 50}
\definecolor{mildblue}{RGB}{50, 50, 200}

\title{CHILL at SemEval-2025 Task 2: You Can’t Just Throw Entities and Hope — Make Your LLM to Get Them Right}

\author{Jaebok Lee \and Yonghyun Ryu \and Seongmin Park \and Yoonjung Choi \\
        Samsung Research, Seoul, South Korea \\
        \texttt{\{jaebok44.lee, yonghyun.ryu, ulgal.park, yj0807.choi\}@samsung.com} \\
        }

\begin{document}
\maketitle
\begin{abstract}
In this paper, we describe our approach for the SemEval 2025 Task 2 on Entity-Aware Machine Translation (EA-MT).
Our system aims to improve the accuracy of translating named entities by combining two key approaches: Retrieval Augmented Generation (RAG) and iterative self-refinement techniques using Large Language Models (LLMs).
A distinctive feature of our system is its self-evaluation mechanism, where the LLM assesses its own translations based on two key criteria: the accuracy of entity translations and overall translation quality. 
We demonstrate how these methods work together and effectively improve entity handling while maintaining high-quality translations.
\end{abstract}

\section{Introduction}\label{introduction}

Entity-Aware Machine Translation (EA-MT) focuses on accurately translating sentences containing named entities, such as movies, books, food, locations, and people.
This task is particularly important because entity names often carry cultural nuances that need to be preserved in the translation process \cite{conia-etal-2024-towards}.
The challenge lies in the fact that named entities typically cannot be translated through simple word-for-word or literal translations.
For instance, the movie "Night at the Museum" is translated as "박물관이 살아있다" (meaning "The Museum is Alive") in Korean, rather than the literal translation "박물관에서의 밤."
This example illustrates why literal translations of entity names can be inappropriate.
This issue becomes particularly critical in domains such as journalism, legal, or medical contexts, where incorrect entity translations can significantly compromise factual accuracy and trustworthiness.

The SemEval 2025 Task 2 \cite{conia2025semeval} addresses this challenge by requiring participants to develop systems that correctly identify named entities and transform them into their appropriate target-language forms.
In this paper, we present our system, which combines Retrieval-Augmented Generation (RAG) and self-refine \cite{madaan2024self} approaches.
Our system first retrieves entity information (labels and descriptions) from Wikidata \cite{wikidata} IDs provided by the task organizers.
This information is then incorporated into prompts for a Large Language Model (LLM).
However, we discovered that merely providing entity information to the LLM does not guarantee accurate entity-aware translations.
To address this limitation, we implemented the self-refine framework where the same LLM model evaluates the initial translation based on two criteria: entity label accuracy and overall translation quality.
In summary, our approach integrates both RAG and self-refine frameworks to achieve optimal translation results.
We also present case studies demonstrating concrete improvements achieved through our feedback mechanism.

\section{Related Work}\label{related_work}

\subsection{Entities in Knowledge Graph}

Knowledge graph component retrieval from source texts has been extensively studied through various approaches, including dense retrieval \cite{conia-etal-2024-towards,karpukhin-etal-2020-dense,wu-etal-2020-scalable,li-etal-2020-efficient} and constrained decoding \cite{genre,GenRL,lee2024sparkle}.
These retrieved information has been successfully applied to various tasks, such as question answering and machine translation.
In the context of machine translation, several studies have focused on entity name transliteration, converting entity names from one script to another \cite{sadamitsu-etal-2016-name,ugawa-etal-2018-neural,zeng-etal-2023-extract}.
However, these studies did not address the transcreation of entity names.
Another line of research has explored improving MT models by augmenting training datasets to enhance entity coverage \cite{hu-etal-2022-deep,liu-etal-2021-mulda}.
While our approach directly utilizes gold Wikidata IDs and retrieves related information, it can be potentially combined with the aforementioned retrieval methods.

\subsection{Retrieval Augmented Generation}

Retrieval Augmented Generation (RAG) has been widely adopted to enhance machine translation accuracy. \citet{zhang-etal-2018-guiding} demonstrated its effectiveness in improving MT quality, particularly for low-frequency words. \citet{bulte-tezcan-2019-neural} employed fuzzy matching techniques for dataset augmentation. More recently, \citet{conia-etal-2024-towards} leveraged RAG with Large Language Models to achieve cross-cultural machine translation.

\subsection{Post Editing and Refinement}

Post-editing has become a crucial part of machine translation workflows to enhance initial translation quality.
Large Language Models (LLMs) have emerged as particularly effective tools for providing translation feedback.
\citet{raunak-etal-2023-leveraging} employed GPT-4 to generate feedback and post-edit Microsoft Translator outputs, while \citet{kocmi-federmann-2023-gemba} utilized GPT-4 to provide MQM-style feedback for machine translation results.
The concept of self-refinement, introduced by \citet{madaan2024self}, allows Large Language Models to generate outputs, feedback and iterate by itself, leading to improved performance across various tasks.
In our work, we adapt this self-refinement framework for entity-aware machine translation, as we found that RAG alone is insufficient for accurate translation.

\section{Dataset}

\begin{table}[h]
    \centering
    \begin{tabular}{lccc}
        \toprule
        Language & Train & Valid & Test \\
        \midrule
        Italian & 3,739 & 730 & 5,097 \\
        Spanish & 5,160 & 739 & 5,337 \\
        French & 5,531 & 724 & 5,464 \\
        German & 4,087 & 731 & 5,875 \\
        Arabic & 7,220 & 722 & 4,546 \\
        Japanese & 7,225 & 723 & 5,107 \\
        Chinese & - & 722 & 5,181 \\
        Korean & - & 745 & 5,081 \\
        Thai & - & 710 & 3,446 \\
        Turkish & - & 732 & 4,472 \\
        \midrule
        Total & 32,962 & 7,278 & 49,606 \\
        \bottomrule
    \end{tabular}
    \caption{Dataset distribution across languages}
    \label{tab:language_data}
\end{table}

The dataset for this task was provided by the organizers in multiple stages: sample, training, validation, and test sets as shown in Table \ref{tab:language_data}.
The sample data served as an initial reference to demonstrate the format and task requirements.
Each dataset, except for the test set, contains English source texts paired with translations in ten target languages: Italian, Spanish, French, German, Arabic, Japanese, Chinese, Korean, Thai, and Turkish.
A typical data entry consists of an English sentence, at least one corresponding translation in a target language, and an associated Wikidata ID for reference.
For instance, the English question ``What year was The Great Gatsby published?'' is paired with its Korean translation ``위대한 개츠비는 몇 년도에 출판되었나요?'' and linked to the Wikidata  ID Q214371.
The test set, which contained approximately 5,000 sentences for each language direction (totaling 49,606 sentences), was released without ground-truth target references.
The official evaluation was conducted using withheld references that were later made available by the organizers.

\section{System Description}

This section provides a detailed description of our system for entity-aware machine translation.
Our approach combines Retrieval-Augmented Generation (RAG) with self-refinement to ensure accurate entity labeling and high-quality translation.

\subsection{Retrieval-Augmented Generation}

\begin{listing}[t]
    \begin{minted}[frame=lines,
                    framesep=3mm,
                    linenos=false,
                    bgcolor=LightGray,
                    fontsize=\tiny,
                    breaklines=true,
                    tabsize=2]{text}
    \end{minted}
    \caption{Prompt template for initial translation}
    \label{listing:init}
\end{listing}

Our system utilizes the gold entity (Wikidata ID) provided in the test dataset.
We begin by extracting the entity labels, which are essential for accurate translation of entity names.
Additionally, we incorporate entity descriptions, as they play a vital role in entity identification and context understanding \cite{genre,wu-etal-2020-scalable}.
These descriptions help the model distinguish between different entity types and generate contextually appropriate translations.
As illustrated in Example \ref{listing:init}, we embed the entity information within the prompt, instructing the model to consider this information when generating the translation. 
The entity information is retrieved using the Wikidata REST API\footnote{https://www.wikidata.org/w/rest.php/wikibase/v1}.

Formally, given a source text $x$, prompt $p_{gen}$, entity information $e$, and model $M$, the initial translation $y_0$ is generated as:

\begin{equation}
    y_0 = M(p_{gen}||e||x)
\end{equation}

\subsection{Self-Refine}

\begin{listing}[t]
    \begin{minted}[frame=lines,
                    framesep=3mm,
                    linenos=false,
                    bgcolor=LightGray,
                    fontsize=\tiny,
                    breaklines=true,
                    tabsize=2]{text}
    \end{minted}
    \caption{Prompt template for feedback}
    \label{listing:feedback}
\end{listing}

After generating the initial translation, we implement an iterative refinement process to enhance the output quality.
Given a feedback prompt $p_{fb}$ and a generated translation $y_t$, the process begins with the model generating self-feedback:

\begin{equation}
    fb_t = M(p_{fb}||e||x||y_t)
\end{equation}

As shown in Example \ref{listing:feedback}, the model evaluates with two key criteria: the accuracy of entity label translation and the grammatical correctness of the translation.
Both criteria are weighted equally, 5 points each—total 10, to align with the task's evaluation metric, which uses the mean of M-ETA and COMET.
To ensure structured feedback, we created few-shot examples across languages, which will be described in Section \ref{sec:few}.

Given a refine prompt $p_{rf}$ and a feedback history, the refinement process alternates between feedback and improvement steps.

\begin{equation}
\label{eq:refine}
    y_{t+1} = M(p_{rf}||e||x||y_0||fb_0||...||y_t||fb_t)
\end{equation}

The process terminates when either the feedback achieves a perfect score of 10 or reaches the maximum number of iterations.
Because each feedback–refinement cycle requires two additional LLM calls (one to generate feedback and one to revise the translation), the computational cost grows linearly with the number of iterations and should therefore be carefully considered.
We set the maximum number of trials to 2 due to budget constraints.

As shown in Equation \ref{eq:refine}, the model incorporates the history of previous translations and their feedback, enabling it to learn from past mistakes and improve both accuracy and overall translation quality.
For detailed prompt, refer to Appendix \ref{sec:iterate}.

\setlength{\dblfloatsep}{1.5em}  

\begin{table*}[t!]
    \centering
    \begin{tabular}{lcccccccccc}
        \toprule
        \textbf{Method} & \textbf{AR} & \textbf{DE} & \textbf{ES} & \textbf{FR} & \textbf{IT} & \textbf{JA} & \textbf{KO} & \textbf{TH} & \textbf{TR} & \textbf{ZH} \\
        \midrule
        \textbf{GPT-4o} & 56.54 & 57.86 & 62.32 & 55.49 & 58.52 & 56.15 & 49.28 & 33.44 & 56.98 & 48.89 \\
        \textbf{  +RAG} & 92.75 & 88.94 & 92.18 & 91.44 & 93.54 & 92.02 & 91.94 & 91.72 & 88.27 & 84.68 \\
        \textbf{  +Refine} & 93.03 & 89.43 & 92.37 & 91.71 & 94.01 & 93.17 & 92.98 & 92.87 & 89.93 & 85.06 \\
        \bottomrule
    \end{tabular}
    \caption{Results across languages with the harmonic mean of M-ETA and Comet scores. Language codes: Arabic (AR), German(DE), Spanish (ES), French (FR), Italian (IT), Japanese (JA), Korean (KO), Thai (TH), Turkish (TR), and Chinese(ZH). For the per‑metric results, refer to Appendix \ref{sec:detailed_result}.}
    \label{tab:method_comparison}
\end{table*}

\begin{table*}[h]
    \begin{tcolorbox}
    \small
    \centering
    \begin{tabularx}{\textwidth}{ X }
        \addlinespace[0.1cm]
        (Entity Feedback) \tabularnewline
        \textbf{Source:} ``When was \textcolor{mildblue}{White Army, Black Baron} first performed?"\tabularnewline
        \textbf{Init:} ``\textcolor{mildblue}{백군, 흑남작}이 처음 공연된 것은 언제인가요?"\tabularnewline
        \textcolor{mildred}{\textbf{Feedback:}} ``...the reference Korean label for this entity is '붉은 군대는 가장 강력하다,' which is a more established and accurate translation of the song's title in Korean" \tabularnewline
        \textcolor{mildred}{\textbf{Refined:}} ``붉은 군대는 가장 강력하다가 처음 연주된 것은 언제인가요?" \tabularnewline
        
        \addlinespace[0.2cm]

        (Translation Feedback) \tabularnewline
        \textbf{Source:} ``Can you recommend any \textcolor{mildblue}{similar webcomics} to Please Take My Brother Away!?"\tabularnewline
        \textbf{Init:} ``비슷한 웹툰으로 오빠를 고칠 약은 없어!\textcolor{mildblue}{를 추천}해 주실 수 있나요?"\tabularnewline
        \textcolor{mildred}{\textbf{Feedback:}} ``The original asks for recommendations of *similar* webcomics, but the translation asks if '오빠를 고칠 약은 없어!' itself can be recommended ..." \tabularnewline
        \textcolor{mildred}{\textbf{Refined:}} ``오빠를 고칠 약은 없어!와 비슷한 웹툰을 추천해 주실 수 있나요?" \tabularnewline
    \end{tabularx}
    \end{tcolorbox}
    \caption{Case study of feedback and refinement}
    \label{tab:case-study}
\end{table*}

\section{Experimental Setup}
\label{sec:setup}

\subsection{Model and Inference}

Our system employs the GPT-4o model as the primary translation and feedback generator.
We used the model without any fine-tuning, relying solely on prompt engineering to achieve the desired results.

\subsection{Few-shot Example Generation}
\label{sec:few}

For the feedback and iteration prompts, we carefully crafted few-shot examples to guide the model's behavior.
The process of creating these examples varied by language.
Being native Korean speakers, we manually created examples by deliberately introducing errors into reference translations.
This allowed us to demonstrate various types of translation errors and appropriate feedback.
We then leveraged GPT-4o to generate examples for the remaining nine language pairs, using our Korean examples as templates.
This ensured consistency in the feedback and iteration patterns across all language directions.

\subsection{Evaluation Metrics}

The shared task evaluation combines two metrics using their harmonic mean.
COMET \cite{rei-etal-2020-comet} is a metric based on pretrained language models that evaluates the overall quality of machine translation outputs.
Additionally, M-ETA (Manual Entity Translation Accuracy) \cite{conia-etal-2024-towards} serves as a specialized metric designed to assess translation accuracy specifically at the entity level.
This combination of metrics ensures that both general translation quality and entity-specific accuracy are considered in the final evaluation.

\section{Results and Analysis}

\subsection{Overall Performance}

Table \ref{tab:method_comparison} presents a comprehensive analysis of our system's performance across different configurations and language pairs on the test dataset.
In the baseline GPT-4o model, we use a basic translation prompt suggested by \citet{xu2024a}.
Despite the source texts being simple and concise questions, the baseline GPT-4o model demonstrates relatively poor performance across different language pairs.
This is primarily due to its inaccuracy in translating entity labels, which results in a lower M-ETA score.

The application of Retrieval-Augmented Generation (RAG) leads to substantial performance improvements across all language pairs.
This significant enhancement is attributed to our utilization of oracle Wikidata IDs from the dataset, from which we extract precise entity labels and descriptions.
Our results demonstrate Large Language Model's capability to successfully incorporate the provided entity information into accurate translations.

The addition of the self-refinement process further enhances the translation quality, albeit with more modest improvements.
We observe consistent performance gains across all language directions, with improvements ranging from 0.19 to 1.66 \%p.
These results validate the effectiveness of both RAG approach and self-refinement mechanism in the context of entity-aware machine translation.

\subsection{Case Study}

Our feedback prompt incorporated two primary evaluation criteria: entity name accuracy and translation quality.
To validate the model's ability to effectively evaluate these criteria, we present two representative cases where improvements were observed in either entity naming or translation quality, as shown in Table \ref{tab:case-study}.

In the first case (Entity Feedback), we observe how the model handles entity name translation.
The initial translation attempted a literal, word-for-word approach, translating ``White Army'' and ``Black Baron'' directly into their Korean equivalents ``백군'' and ``흑남작''.
The feedback procedure identified this literal translation as inadequate, noting that the established Korean title for this entity is ``붉은 군대는 가장 강력하다''.
This case demonstrates that merely providing entity information in the prompt is insufficient for accurate translation; the model requires explicit feedback to generate the correct entity labels.

The second case (Translation Feedback) illustrates the model's ability to correct contextual misunderstandings.
The initial translation misinterpreted the source text's intent, transforming a request for ``finding similar webcomics'' into a request for ``recommending the webcomic itself''.
Through the feedback process, the model recognized this semantic error and generated a refined translation that accurately conveyed the original meaning, asking for recommendations of webcomics similar to the referenced webcomic.
This example highlights the effectiveness of our feedback mechanism in improving not just lexical accuracy but also semantic coherence.

\begin{table}[t!]
\centering
\begin{tabular}{lcc}
\hline
\textbf{Lang} & \textbf{$\rho$} & \textbf{$r$} \\
\hline
DE & 0.17 & 0.21 \\
ES & 0.08 & 0.12 \\
FR & 0.07 & 0.11 \\
IT & 0.03 & 0.07 \\
\hline
\end{tabular}
\caption{Correlation between Levenshtein edit distance ratio and M-ETA scores, measured using Spearman's rank correlation coefficient ($\rho$) and Point-Biserial correlation coefficient ($r$).}
\label{tab:impact}
\end{table}

\subsection{Label Similarity and Accuracy}

We additionally investigated whether the similarity between an entity's English label and its foreign label influences translation accuracy (M-ETA score).
Our hypothesis was that greater differences between entity labels might negatively impact the LLM's translation performance.
To measure label similarity, we used the Levenshtein edit distance ratio.
We analyzed the relationship using two correlation metrics: Spearman's rank correlation coefficient and the Point-Biserial correlation coefficient.
These metrics were chosen for their suitability in analyzing relationships between a continuous variable (edit distance ratio) and a binary outcome (correct/incorrect translation).

We limited our analysis to languages using the Latin script (German, Spanish, French, and Italian) as other languages would consistently yield edit distance ratios approaching 1.
As shown in Table \ref{tab:impact}, the correlation coefficients are consistently low across all languages.
These results suggest that the similarity between entity labels in English and foreign languages has little impact on translation accuracy, indicating that other factors likely play more significant roles in determining translation success.

\section{Conclusion}

In this paper, we present an effective approach to Entity-Aware Machine Translation that combines Retrieval-Augmented Generation (RAG) with a self-refinement mechanism.
Our system features a two-criteria feedback system that identifies and corrects both entity label inaccuracies and translation errors.
When tested against the baseline GPT-4o model, our system demonstrates significant improvements across all language pairs in the task dataset.
The experimental results highlight two key findings:
(i) the integration of RAG with entity information from external knowledge substantially improves translation accuracy.
(ii) self-refinement mechanism consistently enhances translation quality across all language pairs through iterative feedback and correction.
Our case studies reveal that the system effectively addresses both entity-specific challenges and general translation issues.
These results suggest that combining knowledge retrieval with self-refinement is a promising direction for entity-aware machine translation.
Looking ahead, future work could explore incorporating entity retrieval methods without using gold entity.

\bibliography{anthology,custom}

\begin{thebibliography}{21}
\providecommand{\natexlab}[1]{#1}

\bibitem[{Bulte and Tezcan(2019)}]{bulte-tezcan-2019-neural}
Bram Bulte and Arda Tezcan. 2019.
\newblock \href {https://doi.org/10.18653/v1/P19-1175} {Neural fuzzy repair: Integrating fuzzy matches into neural machine translation}.
\newblock In \emph{Proceedings of the 57th Annual Meeting of the Association for Computational Linguistics}, pages 1800--1809, Florence, Italy. Association for Computational Linguistics.

\bibitem[{Conia et~al.(2024)Conia, Lee, Li, Minhas, Potdar, and Li}]{conia-etal-2024-towards}
Simone Conia, Daniel Lee, Min Li, Umar~Farooq Minhas, Saloni Potdar, and Yunyao Li. 2024.
\newblock \href {https://doi.org/10.18653/v1/2024.emnlp-main.914} {Towards cross-cultural machine translation with retrieval-augmented generation from multilingual knowledge graphs}.
\newblock In \emph{Proceedings of the 2024 Conference on Empirical Methods in Natural Language Processing}, pages 16343--16360, Miami, Florida, USA. Association for Computational Linguistics.

\bibitem[{Conia et~al.(2025)Conia, Li, Navigli, and Potdar}]{conia2025semeval}
Simone Conia, Min Li, Roberto Navigli, and Saloni Potdar. 2025.
\newblock Semeval-2025 task 2: Entity-aware machine translation.
\newblock \emph{In Proceedings of the 19th International Workshop on Semantic Evaluation (SemEval2025)}.

\bibitem[{{De Cao} et~al.(2021){De Cao}, Izacard, Riedel, and Petroni}]{genre}
Nicola {De Cao}, Gautier Izacard, Sebastian Riedel, and Fabio Petroni. 2021.
\newblock \href {https://openreview.net/forum?id=5k8F6UU39V} {Autoregressive entity retrieval}.
\newblock In \emph{9th International Conference on Learning Representations, {ICLR} 2021, Virtual Event, Austria, May 3-7, 2021}. OpenReview.net.

\bibitem[{Hu et~al.(2022)Hu, Hayashi, Cho, and Neubig}]{hu-etal-2022-deep}
Junjie Hu, Hiroaki Hayashi, Kyunghyun Cho, and Graham Neubig. 2022.
\newblock \href {https://doi.org/10.18653/v1/2022.acl-long.123} {{DEEP}: {DE}noising entity pre-training for neural machine translation}.
\newblock In \emph{Proceedings of the 60th Annual Meeting of the Association for Computational Linguistics (Volume 1: Long Papers)}, pages 1753--1766, Dublin, Ireland. Association for Computational Linguistics.

\bibitem[{Karpukhin et~al.(2020)Karpukhin, Oguz, Min, Lewis, Wu, Edunov, Chen, and Yih}]{karpukhin-etal-2020-dense}
Vladimir Karpukhin, Barlas Oguz, Sewon Min, Patrick Lewis, Ledell Wu, Sergey Edunov, Danqi Chen, and Wen-tau Yih. 2020.
\newblock \href {https://doi.org/10.18653/v1/2020.emnlp-main.550} {Dense passage retrieval for open-domain question answering}.
\newblock In \emph{Proceedings of the 2020 Conference on Empirical Methods in Natural Language Processing (EMNLP)}, pages 6769--6781, Online. Association for Computational Linguistics.

\bibitem[{Kocmi and Federmann(2023)}]{kocmi-federmann-2023-gemba}
Tom Kocmi and Christian Federmann. 2023.
\newblock \href {https://doi.org/10.18653/v1/2023.wmt-1.64} {{GEMBA}-{MQM}: Detecting translation quality error spans with {GPT}-4}.
\newblock In \emph{Proceedings of the Eighth Conference on Machine Translation}, pages 768--775, Singapore. Association for Computational Linguistics.

\bibitem[{Lee and Shin(2024)}]{lee2024sparkle}
Jaebok Lee and Hyeonjeong Shin. 2024.
\newblock Sparkle: Enhancing sparql generation with direct kg integration in decoding.
\newblock \emph{arXiv preprint arXiv:2407.01626}.

\bibitem[{Li et~al.(2020)Li, Min, Iyer, Mehdad, and Yih}]{li-etal-2020-efficient}
Belinda~Z. Li, Sewon Min, Srinivasan Iyer, Yashar Mehdad, and Wen-tau Yih. 2020.
\newblock \href {https://doi.org/10.18653/v1/2020.emnlp-main.522} {Efficient one-pass end-to-end entity linking for questions}.
\newblock In \emph{Proceedings of the 2020 Conference on Empirical Methods in Natural Language Processing (EMNLP)}, pages 6433--6441, Online. Association for Computational Linguistics.

\bibitem[{Liu et~al.(2021)Liu, Ding, Bing, Joty, Si, and Miao}]{liu-etal-2021-mulda}
Linlin Liu, Bosheng Ding, Lidong Bing, Shafiq Joty, Luo Si, and Chunyan Miao. 2021.
\newblock \href {https://doi.org/10.18653/v1/2021.acl-long.453} {{M}ul{DA}: A multilingual data augmentation framework for low-resource cross-lingual {NER}}.
\newblock In \emph{Proceedings of the 59th Annual Meeting of the Association for Computational Linguistics and the 11th International Joint Conference on Natural Language Processing (Volume 1: Long Papers)}, pages 5834--5846, Online. Association for Computational Linguistics.

\bibitem[{Madaan et~al.(2024)Madaan, Tandon, Gupta, Hallinan, Gao, Wiegreffe, Alon, Dziri, Prabhumoye, Yang et~al.}]{madaan2024self}
Aman Madaan, Niket Tandon, Prakhar Gupta, Skyler Hallinan, Luyu Gao, Sarah Wiegreffe, Uri Alon, Nouha Dziri, Shrimai Prabhumoye, Yiming Yang, et~al. 2024.
\newblock Self-refine: Iterative refinement with self-feedback.
\newblock \emph{Advances in Neural Information Processing Systems}, 36.

\bibitem[{Raunak et~al.(2023)Raunak, Sharaf, Wang, Awadalla, and Menezes}]{raunak-etal-2023-leveraging}
Vikas Raunak, Amr Sharaf, Yiren Wang, Hany Awadalla, and Arul Menezes. 2023.
\newblock \href {https://doi.org/10.18653/v1/2023.findings-emnlp.804} {Leveraging {GPT}-4 for automatic translation post-editing}.
\newblock In \emph{Findings of the Association for Computational Linguistics: EMNLP 2023}, pages 12009--12024, Singapore. Association for Computational Linguistics.

\bibitem[{Rei et~al.(2020)Rei, Stewart, Farinha, and Lavie}]{rei-etal-2020-comet}
Ricardo Rei, Craig Stewart, Ana~C Farinha, and Alon Lavie. 2020.
\newblock \href {https://doi.org/10.18653/v1/2020.emnlp-main.213} {{COMET}: A neural framework for {MT} evaluation}.
\newblock In \emph{Proceedings of the 2020 Conference on Empirical Methods in Natural Language Processing (EMNLP)}, pages 2685--2702, Online. Association for Computational Linguistics.

\bibitem[{Rossiello et~al.(2021)Rossiello, Mihindukulasooriya, Abdelaziz, Bornea, Gliozzo, Naseem, and Kapanipathi}]{GenRL}
Gaetano Rossiello, Nandana Mihindukulasooriya, Ibrahim Abdelaziz, Mihaela Bornea, Alfio Gliozzo, Tahira Naseem, and Pavan Kapanipathi. 2021.
\newblock Generative relation linking for question answering over knowledge bases.
\newblock In \emph{The Semantic Web -- ISWC 2021}, pages 321--337, Cham. Springer International Publishing.

\bibitem[{Sadamitsu et~al.(2016)Sadamitsu, Saito, Katayama, Asano, and Matsuo}]{sadamitsu-etal-2016-name}
Kugatsu Sadamitsu, Itsumi Saito, Taichi Katayama, Hisako Asano, and Yoshihiro Matsuo. 2016.
\newblock \href {https://aclanthology.org/L16-1097} {Name translation based on fine-grained named entity recognition in a single language}.
\newblock In \emph{Proceedings of the Tenth International Conference on Language Resources and Evaluation ({LREC}'16)}, pages 613--619, Portoro{\v{z}}, Slovenia. European Language Resources Association (ELRA).

\bibitem[{Ugawa et~al.(2018)Ugawa, Tamura, Ninomiya, Takamura, and Okumura}]{ugawa-etal-2018-neural}
Arata Ugawa, Akihiro Tamura, Takashi Ninomiya, Hiroya Takamura, and Manabu Okumura. 2018.
\newblock \href {https://aclanthology.org/C18-1274} {Neural machine translation incorporating named entity}.
\newblock In \emph{Proceedings of the 27th International Conference on Computational Linguistics}, pages 3240--3250, Santa Fe, New Mexico, USA. Association for Computational Linguistics.

\bibitem[{Vrande{\v{c}}i{\'c} and Kr{\"o}tzsch(2014)}]{wikidata}
Denny Vrande{\v{c}}i{\'c} and Markus Kr{\"o}tzsch. 2014.
\newblock Wikidata: a free collaborative knowledgebase.
\newblock \emph{Communications of the ACM}, 57(10):78--85.

\bibitem[{Wu et~al.(2020)Wu, Petroni, Josifoski, Riedel, and Zettlemoyer}]{wu-etal-2020-scalable}
Ledell Wu, Fabio Petroni, Martin Josifoski, Sebastian Riedel, and Luke Zettlemoyer. 2020.
\newblock \href {https://doi.org/10.18653/v1/2020.emnlp-main.519} {Scalable zero-shot entity linking with dense entity retrieval}.
\newblock In \emph{Proceedings of the 2020 Conference on Empirical Methods in Natural Language Processing (EMNLP)}, pages 6397--6407, Online. Association for Computational Linguistics.

\bibitem[{Xu et~al.(2024)Xu, Kim, Sharaf, and Awadalla}]{xu2024a}
Haoran Xu, Young~Jin Kim, Amr Sharaf, and Hany~Hassan Awadalla. 2024.
\newblock \href {https://openreview.net/forum?id=farT6XXntP} {A paradigm shift in machine translation: Boosting translation performance of large language models}.
\newblock In \emph{The Twelfth International Conference on Learning Representations}.

\bibitem[{Zeng et~al.(2023)Zeng, Wang, Leng, Guo, Xie, Tan, Qin, and Liu}]{zeng-etal-2023-extract}
Zixin Zeng, Rui Wang, Yichong Leng, Junliang Guo, Shufang Xie, Xu~Tan, Tao Qin, and Tie-Yan Liu. 2023.
\newblock \href {https://doi.org/10.18653/v1/2023.findings-acl.107} {Extract and attend: Improving entity translation in neural machine translation}.
\newblock In \emph{Findings of the Association for Computational Linguistics: ACL 2023}, pages 1697--1710, Toronto, Canada. Association for Computational Linguistics.

\bibitem[{Zhang et~al.(2018)Zhang, Utiyama, Sumita, Neubig, and Nakamura}]{zhang-etal-2018-guiding}
Jingyi Zhang, Masao Utiyama, Eiichro Sumita, Graham Neubig, and Satoshi Nakamura. 2018.
\newblock \href {https://doi.org/10.18653/v1/N18-1120} {Guiding neural machine translation with retrieved translation pieces}.
\newblock In \emph{Proceedings of the 2018 Conference of the North {A}merican Chapter of the Association for Computational Linguistics: Human Language Technologies, Volume 1 (Long Papers)}, pages 1325--1335, New Orleans, Louisiana. Association for Computational Linguistics.

\end{thebibliography}

\newpage
\appendix

\section{Iteration Prompt Template}
\label{sec:iterate}

\begin{listing}[h]
    \begin{minted}[frame=lines,
                    framesep=3mm,
                    linenos=false,
                    bgcolor=LightGray,
                    fontsize=\tiny,
                    breaklines=true,
                    tabsize=2]{text}
    \end{minted}
    \label{listing:sample-conversation}
\end{listing}




\section{Detailed Performance Result}
\label{sec:detailed_result}

\begin{table}[H]
\centering
\small
\begin{tabular}{l|c|c|c|c|c|c}
\hline
\multirow{2}{*}{} &
\multicolumn{2}{c|}{\textbf{GPT‑4o}} &
\multicolumn{2}{c|}{\textbf{+RAG}} &
\multicolumn{2}{c}{\textbf{+Refine}} \\
 & \textbf{C} & \textbf{M} & \textbf{C} & \textbf{M} & \textbf{C} & \textbf{M} \\
\hline
AR & 88.80 & 41.48 & 93.34 & 92.17 & 94.23 & 91.86 \\
DE & 88.25 & 43.04 & 92.71 & 85.46 & 94.08 & 85.23 \\
ES & 88.86 & 48.00 & 93.80 & 90.61 & 95.00 & 89.88 \\
FR & 86.40 & 40.88 & 92.28 & 90.61 & 93.54 & 89.95 \\
IT & 87.28 & 44.02 & 94.46 & 92.64 & 95.65 & 92.43 \\
JA & 82.57 & 42.54 & 94.67 & 89.53 & 95.61 & 90.86 \\
KO & 85.20 & 34.67 & 94.22 & 89.77 & 95.21 & 90.85 \\
TH & 72.25 & 21.76 & 92.40 & 91.06 & 94.26 & 91.53 \\
TR & 84.31 & 43.03 & 94.50 & 82.83 & 95.63 & 84.86 \\
ZH & 81.92 & 34.85 & 92.55 & 78.06 & 93.86 & 77.77 \\
\hline
\end{tabular}
\caption{COMET (C) and M‑ETA (M) scores for GPT‑4o alone, with retrieval‑augmented generation (+RAG), and with iterative refinement (+Refine) across languages.}
\label{tab:rag_refine_langs}
\end{table}

\end{document}